\documentclass[cha,
 amsmath,amssymb,
 reprint,%
]{revtex4-1}

\usepackage{graphicx}
\usepackage{dcolumn}
\usepackage{bm}

\usepackage[utf8]{inputenc}
\usepackage[T1]{fontenc}
\usepackage{mathptmx}
\usepackage{etoolbox}
\usepackage{url}
\usepackage{orcidlink} 

\newcommand{\KL}{\text{KL}}

\makeatletter
\def\@email#1#2{%
 \endgroup
 \patchcmd{\titleblock@produce}
  {\frontmatter@RRAPformat}
  {\frontmatter@RRAPformat{\produce@RRAP{*#1\href{mailto:#2}{#2}}}\frontmatter@RRAPformat}
  {}{}
}%
\makeatother
\begin{document}

\preprint{AIP/123-QED}

\title[]{Intrinsic Dimensionality of Fermi-Pasta-Ulam-Tsingou High-Dimensional Trajectories Through Manifold Learning: A Linear Approach}

\author{Gionni Marchetti\,\orcidlink{0000-0002-8045-5878}} \email{gionnimarchetti@gmail.com}
\affiliation{Barcelona, Spain
}%

\date{\today}

\begin{abstract}
A data-driven approach based on unsupervised machine learning is proposed to infer the intrinsic dimensionality of high-dimensional trajectories in the Fermi–Pasta–Ulam–Tsingou (FPUT) model. Principal component analysis (PCA) is applied to trajectory data accurately computed using a symplectic integrator, comprising $n_s = 4,000,000$ data points from the FPUT $\beta$ model with $N = 32$ coupled harmonic oscillators. By estimating the intrinsic dimension $m^{\ast}$ using multiple methods (participation ratio, Kaiser rule, and the Kneedle algorithm), it is found that $m^{\ast}$ increases with the model’s nonlinearity. Interestingly, in the weakly nonlinear regime ($\beta \lesssim 1.1$), for trajectories initialized by exciting the first mode ($k=1$), the participation ratio estimates  $m^{\ast} = 2, 3$,  strongly suggesting that quasi-periodic motion on a low-dimensional Riemannian manifold underlies the characteristic energy recurrences observed in the FPUT model.
\end{abstract}

\maketitle

\begin{quotation}
``\textbf{The Fermi–Pasta–Ulam–Tsingou (FPUT) model aims to simulate the dynamics of a one-dimensional chain of $N$ weakly coupled harmonic oscillators to investigate the system's route to thermalization. According to ergodic theory, one would expect the system’s trajectories to lie on an $(n - 1)$-dimensional hypersurface of constant energy, where $n$ is the dimension of the phase space. However, the energy recurrences observed for weak nonlinearities suggest that the trajectories may instead lie on low-dimensional invariant tori, in accordance with the Kolmogorov–Arnol'd–Moser (KAM) theorem. In this work, we estimate the intrinsic dimensionality of trajectory data, consisting of 4,000,000 points,from the FPUT $\beta$ model with $N = 32$, using a manifold learning approach based on principal component analysis (PCA). We show that the intrinsic dimensionality increases with the nonlinear strength of the model, as characterized by the parameter $\beta$ and the energy density $\epsilon$. Remarkably, for weak nonlinearities, where characteristic energy recurrences are observed, we find that the system's dynamics evolve on a Riemannian manifold with an intrinsic dimension equal to $2$ or $3$, when the initial condition corresponds to excitation of the first mode ($k = 1$).''} 
\end{quotation}

\section{\label{sec:introduction} Introduction}

The Fermi-Pasta-Ulam-Tsingou (FPUT) model was conceived primarily to test the validity of the equipartition theorem, a fundamental result of classical statistical mechanics, through computer simulations of its nonlinear dynamics~\cite{fermi1955, FORD1992, weissert1997, Falcioni2004}. Fermi who ``foresaw the dawning of computational science''~\cite{Dongarra2024}, expected that the simulations of the dynamics of a one-dimensional set of weakly coupled harmonic oscillators obtained through MANIAC-I computer~\cite{Anderson1986, Porter2009}, would support the equipartition theorem, and hence confirm Boltzmann's~\emph{ergodic hypothesis}~\cite{gallavotti2014, liu2021, Moore2015} \footnote{What Boltzmann meant with ergodic hypothesis probably was what is referred to as Ehrenfest's quasi-ergodic hypothesis}. 
Note that the ergodic hypothesis is commonly assumed to hold when carrying out the molecular dynamics simulations~\cite{kuhne2020}, even though many systems, such as glasses and nearly harmonic solids, are not ergodic in principle~\cite{Frenkel2002}. Nevertheless, the mode energy recurrences observed in simulations of the FPUT model, first performed by Mary Tsingou, appeared to challenge this assumption~\cite{fermi1955, TUCK1972, FORD1992, Dauxois2008}. This surprising result, known as the FPUT paradox, prompted numerous efforts to understand the system's dynamics through both numerical and theoretical investigations, leading to several important findings (see, e.g., Refs.~\onlinecite{Izrailev1966,fucito1982,Livi1985,berman2005,penati2007,onorato2015}; this list is by no means exhaustive). In this regard, it is worth recalling here that the Kolmogorov-Arnol'd-Moser (KAM) theorem was proposed as a plausible explanation of the quasi-periodic behavior~\cite{arnold1989, RINK2003, Ford1970}. According to KAM theory, one would expect that at low energy densities or for small nonlinearities, the trajectories are subject to a periodic motion on invariant topological tori embedded in the phase space of dimension $n$ ($n = 2N$, where $N$ is the number of oscillators)~\cite{RINK2003, Masoliver2011,nachiket2024}. 
On the other hand, the state of the FPUT system can be considered a point in the phase space as typically assumed within the microcanonical formalism of statistical mechanics~\cite{huang1987}. As a result, during the time-evolution, such a point traces out a trajectory that always stays on the hypersurface of constant energy $\Sigma_E = \{\left(q,p\right):  H\left(q, p\right) = E \}$, where $H$ and $q,p$ denote the system's Hamiltonian and the canonical coordinates, respectively. Accordingly,  $\Sigma_E$ has dimension $n-1$, but the KAM invariant tori have dimensions $n/2$. 

The abstract geometric framework described above suggests a critical relationship between the intrinsic (or effective) dimensionality of the trajectories in phase space and the nonlinearity of the FPUT model, which depends on the model parameters $\alpha$ and $\beta$, as well as the energy density $\epsilon$ (see Sec.~\ref{FPUT} for details).

In light of this, our objective is to unravel this relationship by investigating the intrinsic dimension of the trajectory data from the FPUT $\beta$ model, where $\alpha = 0$ and $N = 32$, using a data-driven approach. To this end, we shall focus on entire trajectory data, each formed by $n_s = 4,000,000$ data points, accurately obtained by symplectic integration, with the initial condition corresponding to initially excite either the first energy mode ($k=1$) or the second energy mode ($k=2$). These large data sets, generated for  $\beta \in [0.1, 3]$, capture the full range of typical FPUT phenomenology, from energy recurrences to the path toward thermalization, when $k = 1$ (see Ref.~\onlinecite{giordano2006}). 

Consequently, we apply principal component analysis (PCA), a workhorse of unsupervised machine learning (ML) and statistics~\cite{hastie2017, Joliffe2016, Greenacre2022}, to the data under consideration. PCA is a simple and efficient manifold reduction tool; however, its use involves adopting, as  \emph{working hypothesis}, the assumption that the underlying data structure is linear~\cite{Lever2017, meila2024, tenenbaum2000}. This assumption is not necessarily valid, as demonstrated using t-distributed stochastic neighbor embedding ($t$-SNE)~\cite{vandermaaten2008, Kobak2019, Kobak2021, kobak2024}\footnote{In literature, $t$-SNE is commonly referred to as a nonlinear manifold reduction algorithm.}, which shows that early-stage trajectory data forms closed orbits for weak nonlinearities ($k=1$, $\beta \lesssim 1.1$). However, the linear approach predicts a reasonable monotonic relationship between the dimensionality of the data and the non-linear strength of the model, i.e. $\beta\epsilon$. Additionally, in the weakly nonlinear regime ($k=1$, $\beta \lesssim 1.1$),  it provides an estimate of the intrinsic dimension that closely matches the one obtained using the multi-chart flows method, a Riemannian manifold learning technique recently proposed by Yu et al.~\cite{Yu2025} as discussed in more detail below.

According to PCA, we shall estimate the dimensionality of the trajectory, $m^{\ast}$, using three heuristics: the participation ratio (PR)~\cite{Kramer1993, RECANATESI2022}, the Kaiser criterion (KC), also known as the Kaiser–Gutman rule)\cite{Kaiser1960, jolliffe2002pca}, and the identification of an elbow in the reconstruction error curves~\cite{tenenbaum2000, Geron2017}. Furthermore, elbow detection is automated using the Kneedle algorithm (KA)~\cite{ville2011}. All these methods produce the same qualitative monotonic trend for $m^{\ast}$ as a function of $\beta$; however, $D_{\rm PR}$ underestimates the intrinsic dimensionality, particularly as $\beta$ increases.

Although it remains inconclusive which method is more accurate, given their heuristic nature and the underlying linear assumption, it is worth noting that $D_{\rm PR}$ yields $m^{\ast} = 2\text{--}3$ in the weakly non-linear regime. Remarkably, these estimates align with those obtained using the multi-chart flows approach~\cite{Hauberg2025}. These findings strongly support the following picture: at weak nonlinearities, where energy recurrences are observed, the system exhibits quasiperiodic motion on or near a low-dimensional Riemannian manifold. At the other extreme, large intrinsic dimensions ($m^{\ast} = 37\text{--}38$) are observed under strong nonlinearities (that is, as $\beta \to 3$ when $k=1$), when the system approaches thermal equilibrium.

Finally, in Sec.~\ref{conclusions}, we discuss potential directions for overcoming the limitations of this exploratory study.

\section{\label{FPUT} The Fermi-Pasta-Ulam-Tsingou $\beta$ Model} 

The original Fermi-Pasta-Ulam-Tsingou $\beta$ model describes a one-dimensional system of $N$ coupled harmonic oscillators whose Hamiltonian $H\left(q, p\right)$ where $q=\left(q_0, q_1, \cdots, q_N\right)$ and $p=\left(p_0, p_1, \cdots, p_N\right)$, reads~\cite{fermi1955}
\begin{equation}\label{eq:hamiltonian}
\begin{aligned}
H\left(q, p\right) ={} & \frac{1}{2} \sum_{i=1}^{N} p_{i}^{2} + \frac{1}{2} \sum_{i=0}^{N} \left(q_{i+1} - q_i\right)^{2}  \\
      & + \frac{\beta}{4} \sum_{i=0}^{N}  \left(q_{i+1} - q_i\right)^{4} \, .
\end{aligned}
\end{equation}

The nonlinearity of such a model chiefly arises from the parameter $\beta$. But, it can be shown using scaling arguments that the quantity $\beta \epsilon$  determines the degree of nonlinearity ~\cite{BENETTIN2013,benettin2023}.  Here  $\epsilon$ denotes the energy per particle (or energy density), that is, $\epsilon=E/N$,  $E$ being the total energy.

By means of the normal mode coordinates $a_k$ ($k=1, 2, \cdots, N$)~\cite{giordano2006, FORD1992} for which
\begin{equation}\label{eq:variables}
a_k = \sqrt{\frac{2}{N+1} } \sum_{j=0}^{N} q_j \sin\left( \frac{jk\pi}{N+1}\right) \, ,
\end{equation}
and neglecting the terms arising from the cubic and quartic terms in the Hamiltonian~\cite{pace2019}, 
one can express the energy $E_k$  of  normal $k$-th mode as~\cite{fermi1955, giordano2006}
\begin{equation}\label{eq:energy_mode}
E_k = \frac{1}{2} \left[ \Dot{a}_{k}^2  + \omega_{k}^{2}  a_{k}^2\right]  \, .
\end{equation}
where $\omega_{k} = 2 \sin\left(k\pi/2\left(N+1\right)\right)$ is the frequency of the  normal $k$-th mode.  We note in passing that one can assume in good approximation that for weak nonlinearity $E=\sum_{i=1}^{N} E_k$~\cite{reiss2023}.

In the following, we shall limit ourselves to the $\beta$ model, where $\alpha = 0$, that corresponds to a perturbation of strength $\beta$ ($\beta>0$)  of the linear chain of oscillators due to the quartic potential, i.e., the fourth term of Eq.~\ref{eq:hamiltonian}. Furthermore, we shall study the $\beta$-model dynamics assuming fixed boundary conditions, i.e., $q_0= q_{N+1} = 0$.

The typical initial conditions at time $t=0$ are given as by the following formula~\cite{FORD1992, giordano2006} 
\begin{equation}\label{eq:initial_cond}
q_i\left(0\right) = A \sqrt{\frac{2}{N+1} }\sin\left( \frac{ik\pi}{N+1}\right) \, ,
\end{equation}
where $A$ denotes the amplitude. In the following, we shall set $A=10$ according to Ref.~\onlinecite{giordano2006}. In this work, the initial conditions correspond to the first mode  (i.e., $k=1$) 
or the second mode (i.e., $k=2$) being initially excited, as shown in Fig.~\ref{fig:initial_modes}.

We chose the velocity Verlet algorithm~\cite{verlet1967} for integrating the FPUT  model's canonical equations of motion, dictated by the Hamiltonian (Eq.~\ref{eq:hamiltonian})~\footnote{From the Hamiltonian $H$ follows the canonical equations: $\Dot{q} = \partial H/\partial p$, $\Dot{p} = - \partial H/\partial q$.}.  This algorithm is symplectic as required for the problem at hand~\cite{Hairer2006, BENETTIN2011}, and also a second-order method with local and global integration errors that scale as $\mathcal{O}(h^{4})$ and $\mathcal{O}(h^{2})$, respectively, $h$ being the finite size step~\cite{Eastman2016EnergyCA}. 

We tested our numerical simulations against those reported in Ref.~\onlinecite{giordano2006}, for which it was assumed $h=0.05$, finding an excellent agreement.

\begin{figure*}
    \centering
    \begin{minipage}{0.48\textwidth}
        \centering
        \includegraphics[width=\linewidth]{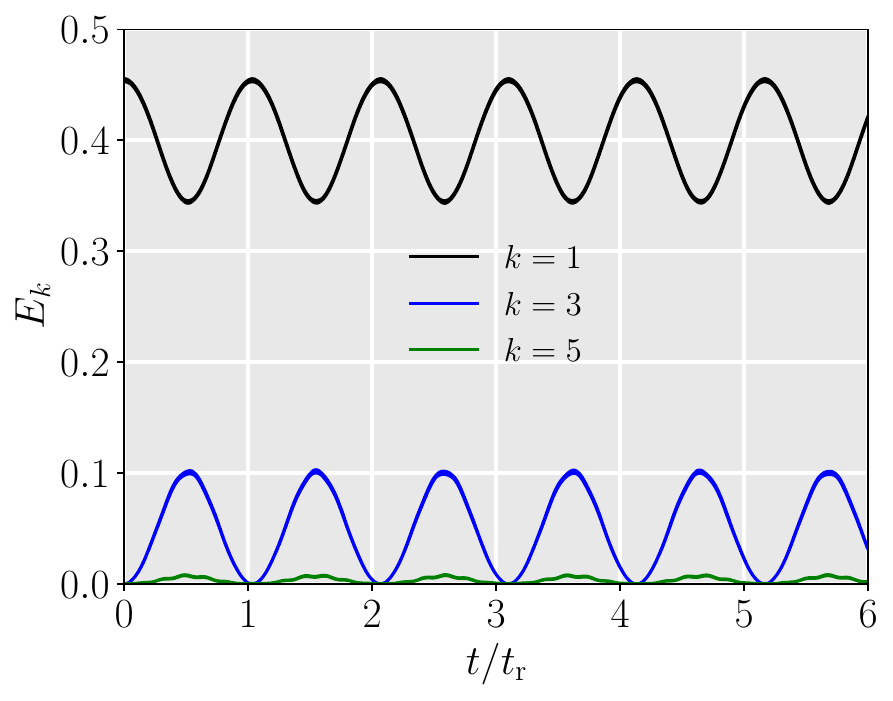}
        \caption{The energy $E_k$  of modes  for $k=1, 3, 5$ as a function of time  $t$ in units of recurrence time $t_r$ ($t_r = 2 \times 10^{5}$) for $\beta$ model with $\beta=0.3$, assuming $N=32$. The system's equations of motion were numerically integrated with size step  $h=0.05$. The initial condition is set to provide the energy $\mathcal{E}_1 \approx 0.45$ to the first normal mode ($k=1, A=10$).
        }
        \label{fig:recurrence}
    \end{minipage}
    \hfill
    \begin{minipage}{0.48\textwidth}
        \centering
        \includegraphics[width=\linewidth]{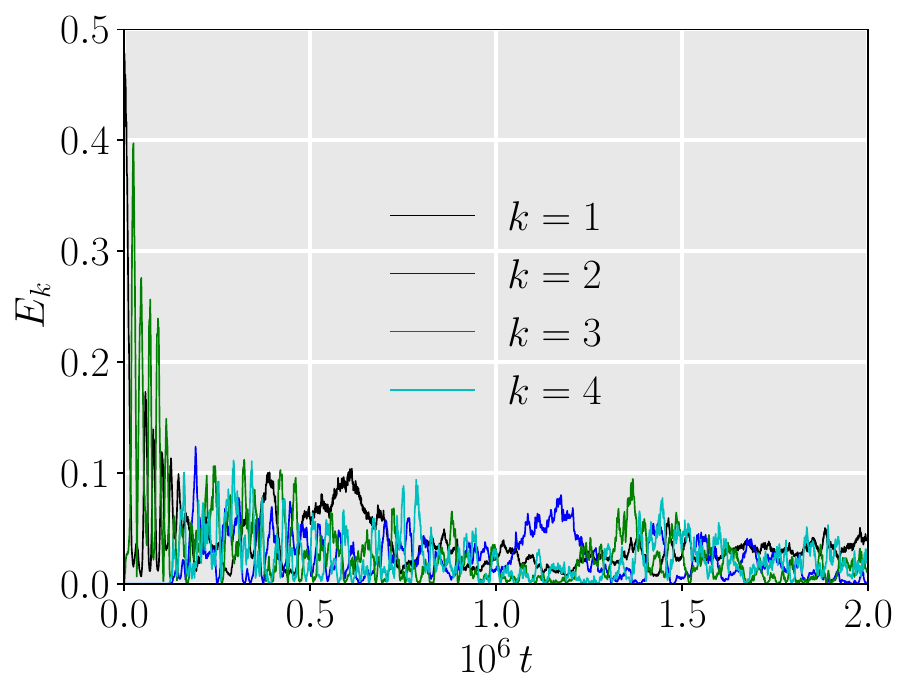}
        \caption{The energy of modes $E_k$ for $k = 1, 2, 3,  4$ as a function of  time  $t$  for $\beta$ model with $\beta=3$, assuming $N=32$. The system's equations of motion were numerically integrated with size step  $h=0.05$. The initial condition is set to provide the energy $\mathcal{E}_1 \approx 0.45$ to the first normal mode ($k=1, A=10$).}
        \label{fig:energy_modes_3.0}
    \end{minipage}
\end{figure*}

In Fig.~\ref{fig:recurrence} we plot the energies $E_k$ for the normal modes  $k=1, 3, 5$ as functions of  time $t$ in units of recurrence time $t_r = 2 \times 10^{5}$~\cite{pace2019}, assuming $\beta=0.3$, and  $N=32$. This initial condition corresponds to initially giving the energy $\mathcal{E}_1$ ($\mathcal{E}_1 \approx 0.45$) to the first normal mode. The time-dependence of these energies $E_k$ illustrates the typical observed energy recurrence phenomenon occurring for small nonlinearities~\cite{TUCK1972}. It is also worth noting that in such a case, there cannot be energy sharing with even modes, that is, modes whose wave number $k$ is equal to an even number. This is due to the symmetric nature of the $\beta$ model~\cite{pace2019, reiss2023}. 
On the other hand, for strong non-linearities, the first mode efficiently shares its energy with the different modes, including the even modes (violation of ``parity conservation'') as shown for the modes $k=1, 2, 3, 4, $ in Fig.~\ref{fig:energy_modes_3.0}, assuming $\beta =3$. In such a case, the system is on a path toward thermalization through irreversible energy sharing among its energy modes.

\section{\label{methods} Methods}

In the following, we shall briefly recall the main results of the unsupervised ML algorithms we employed for the dimensional reduction of the data generated from the high-dimensional FPUT trajectories. We leveraged the principal component analysis to compute the reconstruction error $J_m$ of the original data's orthogonal projections onto a suitable linear subspace $U  \subset  \mathbb{R}^n$ of dimension $m$, while $t$-SNE  helped us visualize in two-dimensions the 
embedding arising from a given trajectory in the early stage of the system's dynamics.

In the present work, a trajectory, including its initial condition, forms a data set $\mathcal{X} = \{x_1, x_2, \cdots, x_{\rm n_s} \}$, where each element $x_i$, is a point in the phase space $\mathbb{R}^n$. Accordingly, each phase point represents the system's position along the orbit as time $t$ increases monotonically from zero.
A $n_s \times n $ data matrix $X$ can be constructed by setting each $x_i$ as a row of  $X$, where $i$ runs from $1$ to $n_s$.

\subsection{PCA and the Reconstruction Error}\label{PCA}

The principal component analysis is a linear unsupervised dimensionality reduction technique~\cite{Joliffe2016, Greenacre2022}, which can be useful for data visualization in a low-dimensional space. PCA finds new uncorrelated variables,  the principal components (PCs), via a linear transformation~\cite{Pearson1901, Hotelling1933}. Accordingly, the axes corresponding to PCs maximally preserve the variance of high-dimensional data in decreasing order. The variances preserved (explained) along the PC axes are the eigenvalues
$\lambda_l$ with $l=1, \cdots, n$ of the (sample) covariance matrix $S$
\begin{equation}\label{eq:covariance}
S =\frac{1}{n_s -1 }  \tilde{X}^T \tilde{X} \, .
\end{equation}
where $\tilde{X}$ is the $n_s \times n$ data matrix $X$, after the standardization procedure of the variables~\cite{Joliffe2016}. As a result, the variables are now scale-free each with zero mean and variance equal to unity, making $S$ a correlation matrix. Note that the mean centering is necessary when the covariance matrix's eigenvalues $\lambda_i$ are computed using the singular value decomposition (SVD)~\cite{ strang1993, stewart1993}.
According to SVD,  $\tilde{X}=WLV^T$ where $W$ and $V$ are two suitable orthogonal matrices, and $L$ is a diagonal matrix~\cite{Joliffe2016, Greenacre2022}. As a result, 
the eigenvalues $\lambda_i$ can be efficiently computed from the equation $\lambda_i = \left(n_s -1\right)^{-1}  s_{i}^{2}$ where $s_i$ are the diagonal entries of $L$. Furthermore, it is assumed that $s_{1}^{2} \geq s_{2}^{2}  \geq \cdots \geq s_{n}^{2} \geq 0 $. In the present work, the singular values $s_i$ of $\tilde{X}$ are computed using the scikit-learn ML library~\cite{pedregosa2011, Geron2017}.

PCA can be understood as an unsupervised ML algorithm that maximally preserves the overall variance of the original high-dimensional data along the principal components~\cite{Pearson1901, Hotelling1933, Joliffe2016} or orthogonally projects the data onto a suitable lower-dimensional linear subspace $U$, commonly known as the principal subspace, of dimension $m$, minimizing
the average reconstruction error $J_m$. Consequently, starting with the data points $x_i$ with $i=1, \cdots, n_s$ in $\mathbb{R}^n$, the reconstruction error $J_m$ to approximate each $x_i$ by its orthogonal projection $\tilde{x}_i \in U$, is the average squared Euclidean distance defined as follows~\cite{Deisenroth2020, hastie2017}
\begin{equation}\label{eq:rec_error}
J_m = \frac{1}{n_s}\sum_{j= 1}^{n_s} \lVert x_j - \tilde{x}_j \rVert_2^{2} \, ,
\end{equation}
where the symbol $\lVert  \cdot \rVert_2$ denotes the Euclidean norm. This error can be computed through the eigenvalue $\lambda_i$, which accounts for the variance preserved by the $i$-th principal component, and reads~\cite{Deisenroth2020}
\begin{equation}\label{eq:rec_error_1}
J_m = \sum_{l= m + 1}^{n} \lambda_l \, .
\end{equation}

Note that Eq.~\ref{eq:rec_error_1} assumes that $ \lambda_1 \ge \lambda_2 \ge \cdots \ge \lambda_i \ge \lambda_{i+1} \ge \cdots \ge \lambda_{n} $~\cite{Deisenroth2020}. Furthermore, the eigenvectors relative to the eigenvalues $\lambda_l$ with $ l \geq m + 1$ constitute the basis of the orthogonal complement of the principal subspace $U$.

\subsection{t-Distributed Stochastic  Neighbor Embedding }\label{TSNE}

In contrast to PCA, $t$-SNE renounces the preservation of the pairwise distances, thereby avoiding the possible issues arising from the high dimensionality of the data. To this end, this algorithm replaces the distances between the data points in $\mathcal{X} = \{x_1, x_2, \cdots, x_{\rm n_s} \}$, where each element $x_i$ belongs to $\mathbb{R}^n$ with a symmetric joint-probability distribution $P$.  Consequently, it searches for a low-dimensional embedding (or map) $\mathcal{Y} = \{y_1, y_2, \cdots, y_{\rm n_s} \}$, characterized by a symmetric joint-probability distribution $Q$, by minimizing, through the gradient descent, an objective function corresponding to the  Kullback-Leibler (KL) divergence $\KL(P\|Q)$ between $P$ and $Q$:
\begin{equation}\label{eq:KL}
	\KL(P\|Q)=\sum_{i=1}^{n_s}  \sum_{j=1, j \neq i }^{n_s}  p_{i j} \log \frac{p_{i j}}{q_{i j}} \, ,
\end{equation}
where the symmetric probabilities $p_{ij} = \left(2 n_s \right)^{-1} \left(p_{i\mid j} + p_{j\mid i}\right)$  and $q_{ij} =  \left(2 n_s \right)^{-1}  \left(q_{i\mid j} + q_{j\mid i}\right)$ depend on the conditional probabilities $p_{j\mid i}$ and $p_{j\mid i}$, respectively. The probabilities $p_{ij}$ and $q_{ij}$ measure the similarity between $x_i$, $x_j$ and $y_i$, $y_j$, respectively. On the other hand, $p_{j\mid i}$ yields the probability that $x_j$ would be a neighbor of $x_i$, as a Gaussian kernel:
\begin{equation}\label{eq:conditional_p}
	p_{j\mid i} = \frac{\exp(- \lVert x_i - x_j \rVert_2^{2} / 2\sigma_i^2)}{\sum_{k=1, k \neq i}^{n_s} \exp(-\lVert x_i - x_k \rVert_2^{2} / 2\sigma_i^2)}  \, ,
\end{equation}
where the width of the kernel $\sigma_i$ represents the local density.  The variance $\sigma_i^2$ is determined by specifying the perplexity parameter $\tau_p$. The latter is assumed to vary from $5$ to $50$, $30$ being the default value~\cite{vandermaaten2008, Kobak2019}. The perplexity can be thought of as the effective number of neighbors.

Similarly, $q_{j\mid i}$ gives the probability that $y_j$ would be a neighbor of $y_i$. However, given a pair of data points belonging to $\mathcal{Y}$, the probability $q_{ij}$ is now based on the t-distribution with one degree of freedom (equivalently, the Cauchy distribution), and reads
\begin{equation}\label{eq:t_student}
	q_{i j}=\frac{\left(1+\left\|y_{i}-y_{j}\right\|_2^{2}\right)^{-1}}{\sum_{k =1, k \neq l}^{n_s}\left(1+\left\|y_{k}-y_{l}\right\|_2^{2}\right)^{-1}}  \, .
\end{equation}

We refer the reader to Ref.~\onlinecite{Kobak2019} for computational details about the implementation of $t$-SNE. In this work, the respective computations will be performed through \textsf{openTSNE}~\cite{policar2024}.

Finally, the Euclidean distance in Eq.~\ref{eq:conditional_p} can be replaced by the cosine distance $d_{\rm cos}$, which is believed to be less affected by high-dimensional data compared to the Euclidean distance~\cite{kobak2024}. The cosine distance reads~\cite{Murphy2012}
\begin{equation}\label{cosine_distance}
d_{\rm cos}\left(x_i, x_j\right) = 1 - \frac{x_i \cdotp x_j}{\lVert x_i   \rVert_2 \lVert  x_j \rVert_2 }  \, .
\end{equation}

\section{\label{discussion} Results and Discussion} 

To begin, we address the limitations of PCA by visualizing two-dimensional embeddings of trajectory data using $t$-SNE, based on the initial condition with $k=1$, $A=10$, and $\beta =0.1, 1.5, 1$. Due to this choice of parameters, characteristic energy recurrences are observed during the dynamics of the model, see analysis in Ref.~\onlinecite{giordano2006}. Accordingly, we consider embeddings of very early-stage entire trajectories corresponding to $n_s =2, 000 \text{--} 10, 000$ data  points~\footnote{Using \textsf{openTSNE},  it is possible to visualize two-dimensional embeddings for large trajectory datasets ($n_s \sim 10^{6}$); however, the emerging patterns are too complex to allow for a clear interpretation}.

\begin{figure*} 
    \includegraphics[width=\textwidth]{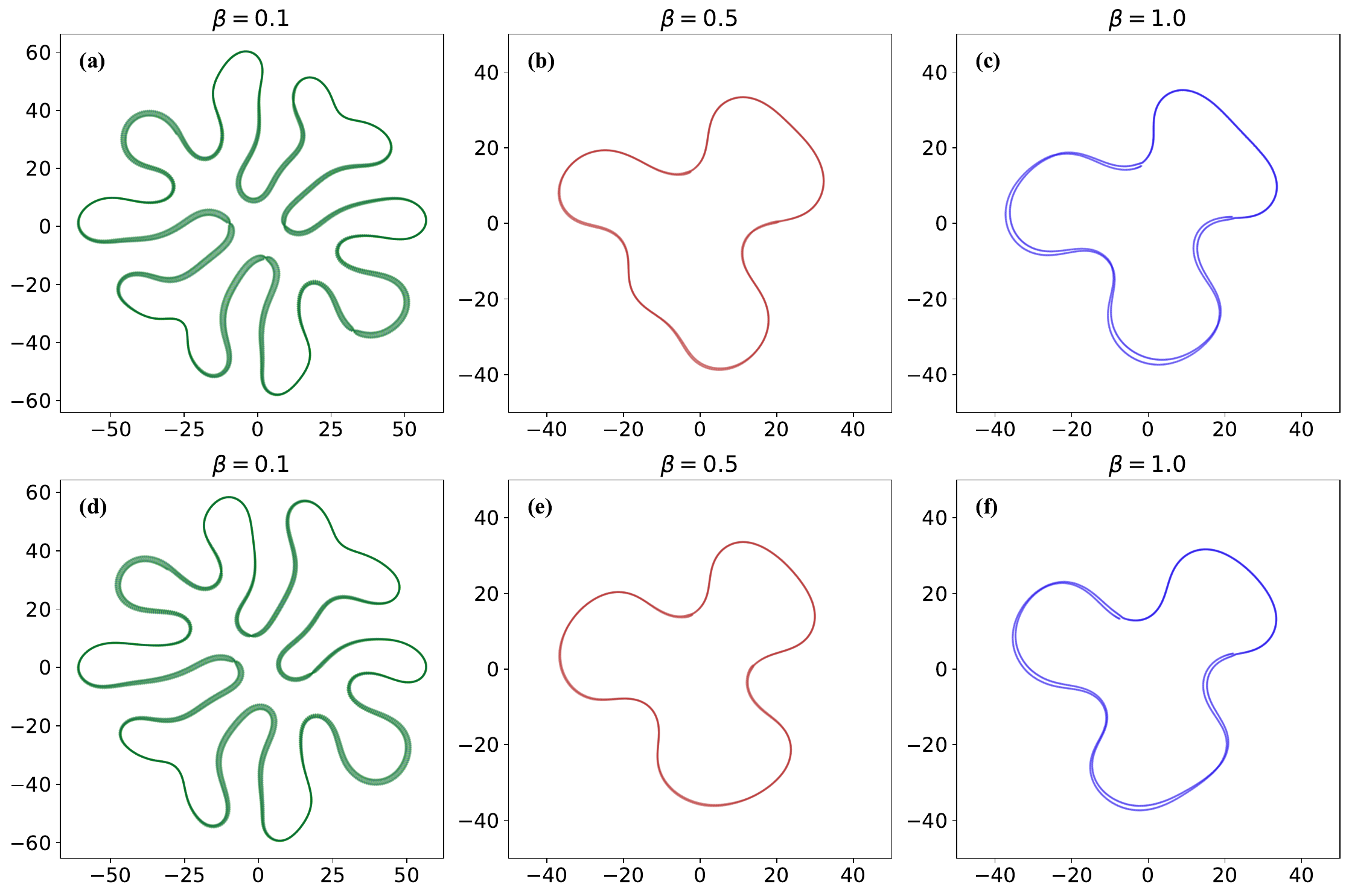}
\caption{$t$-SNE embeddings of the entire trajectories of early-stage dynamics, with $n_s = 10,000$ and $n_s = 2,000$ data points corresponding to $\beta = 0.1$ and $\beta = 0.5, 1$, respectively. The trajectory data were generated with the initial condition $k=1$, $A=10$.  The top panels (a), (b), (c) and bottom panels (d), (e), (f) show embeddings obtained using Euclidean distance and Cosine distance, respectively. PCA initialization was used throughout, and $\tau_p = 50$.}
\label{fig:tsne_small}
\end{figure*}

In Fig.~\ref{fig:tsne_small} the two-dimensional embeddings of trajectory data with  $\beta = 0.1$ and $n_s =10, 000$  (panels (a) and (d)), $\beta = 0.5$ and $n_s =2, 000$  (panels (b) and (e)), and  $\beta = 1$ and $n_s =2, 000$  (panels (e) and (f)) are shown. The $t$-SNE computations were performed setting $\tau_p = 50$, and using the Euclidean distance and the Cosine distance for the embeddings in the top and bottom panels, respectively. It is worth noting that we initialized $t$-SNE using PCA,  this is because only with such an informative initialization can this algorithm preserve both the global and local structures of the data, as recently shown by Kobak and Linderman~\cite{Kobak2021}.
These embeddings reveal that the trajectories form closed orbits, and as a result, the presence of such nonlinear patterns calls into question the use of PCA~\cite{shlens2014, Lever2017}. In this regard,  similar embeddings are obtained using the default perplexity, i.e., $\tau_p =30$ (not shown).
Notably, the negligible differences observed between the embeddings computed using Euclidean and cosine distances strongly suggest that the high dimensionality of the data does not significantly affect the results. Interestingly, the embeddings corresponding to $\beta=0.1$ closely resemble those obtained by applying $t$-SNE to points sampled from a circle with a small amount of Gaussian noise~\cite{Kobak2021}.
Overall, these findings suggest that the data points lie on or near a low-dimensional Riemannian manifold, as demonstrated by the multi-chart flows approach~\cite{Yu2025, Hauberg2025}.

Next, we apply PCA to datasets composed of complete trajectories, consisting of $n_s = 4, 000, 000$ with initial condition  $k=1$,  each generated for values of $\beta$, taken at the fixed step $\Delta \beta = 0.1$ within the interval $[0.1, 3]$. These datasets capture the full range of FPUT dynamics, from energy recurrences to energy sharing among modes as the system approaches thermal equilibrium~\cite{giordano2006}. In contrast, trajectory data initialized with $k=2$ show energy recurrences only when
$\beta=0.1$ (see Fig.~\ref{fig:energy_modes_0.1_k_2}). This fact is a direct consequence of the higher energy density present in the system as explained in Sec.~\ref{fput_facts}. As a result, after an initial transient period (which becomes shorter as $\beta$ increases), the initially excited mode begins to share its energy with the other modes, see Figs.~\ref{fig:energy_modes_0.2_k_2}, ~\ref{fig:energy_modes_0.3_k_2}, ~\ref{fig:energy_modes_0.4_k_2}. 

In the context of PCA, determining the intrinsic dimensionality of the trajectories is equivalent to deciding how many principal components to retain. This is a challenging problem, and it is therefore not surprising that various methods have been proposed. To our knowledge, existing approaches include the Gavish–Donoho optimal hard threshold~\cite{Gavish2014}, the Wachter method~\cite{wachter1976, bai1999, johnstone2001}, the participation ratio~\cite{Kramer1993}, the Kaiser criterion (also known as the Kaiser–Gutman rule)~\cite{Kaiser1960, jolliffe2002pca}, and the identification of the elbow in reconstruction error curves~\cite{tenenbaum2000, Geron2017}.

The Gavish–Donoho optimal hard threshold and Wachter methods are based on random matrix theory~\cite{livan2018random}. Consequently, the transpose of the correlation matrix (see Eq.~\ref{eq:covariance}) is interpreted as a random matrix. Its eigenvalues $\lambda_i$ are compared with those predicted by the Marchenko–Pastur (MP) distribution~\cite{marcenko1967}, in order to identify and discard those that are likely to arise from the white noise. However, we cannot apply these approaches in our case, because the aspect ratio of the data matrix $X$, given by $n/n_s$, is essentially zero ($n/n_s \approx 1.6 \times 10^{-5}$). For the MP distribution to be applicable, the aspect ratio is expected to satisfy $0 < n/n_s  \leq 1$. Furthermore, a very small aspect ratio causes the MP distribution to sharply peak, which poses challenges for accurate numerical integration.

The standard method for estimating the intrinsic dimension $m^{\ast}$ from a reconstruction error curve involves visually identifying the elbow (or equivalently the knee)  of such a curve, beyond which $J_m$ no longer decreases significantly as $m$ increases~\cite{tenenbaum2000, Geron2017}~\footnote{Elbows (knees) can appear in curves with negative (positive) concavity. Elbows typically appear in scree plots, which display the explained variance as a function of the principal components.}. 

\begin{figure*}
    \centering
    \begin{minipage}{0.48\textwidth}
        \centering
        \includegraphics[width=\linewidth]{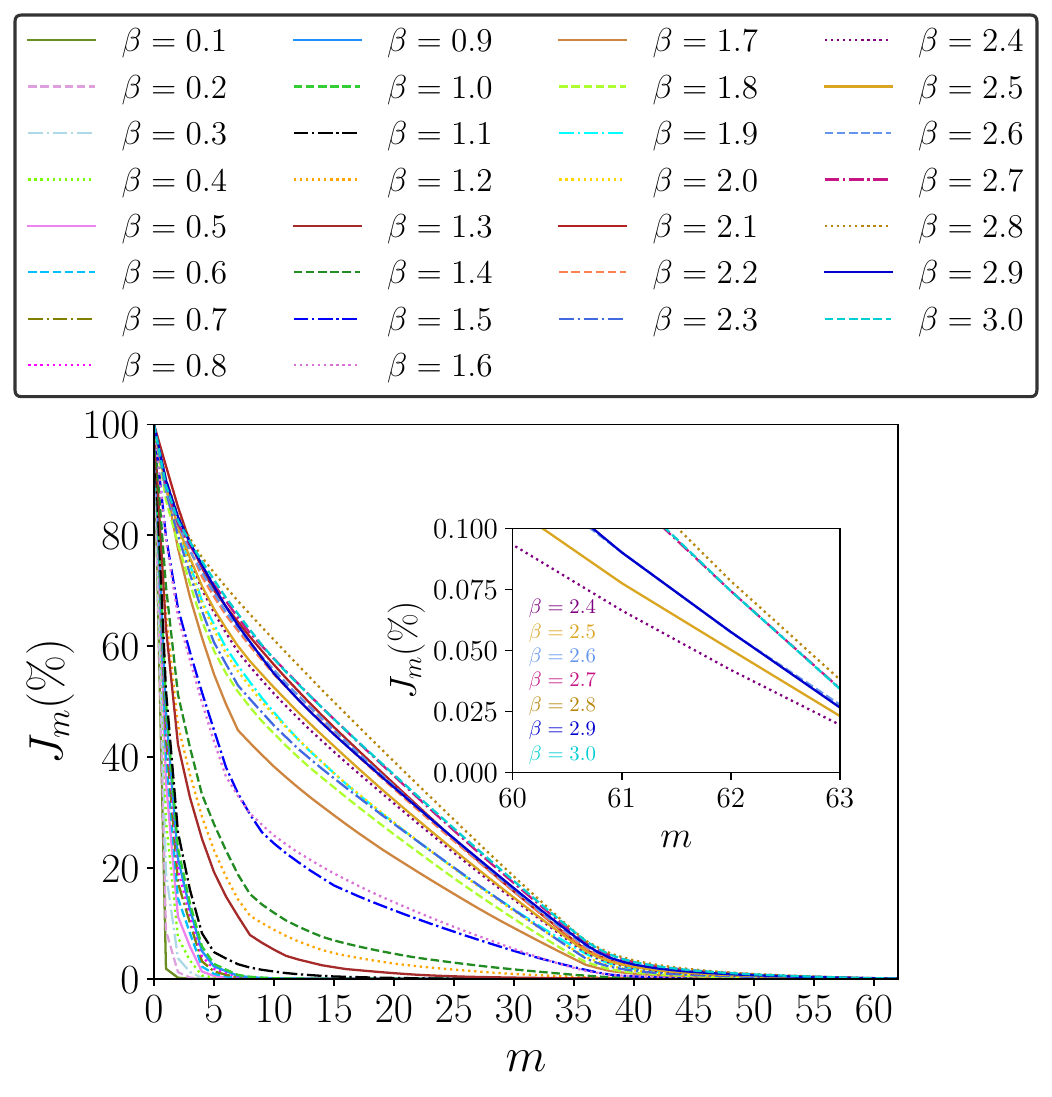}
        \caption{Reconstruction error $J_m$ in percentage ($\%$) as a function of the dimension $m$ of the best-fitting subspace $U$ for $\beta \in [0.1, 3]$, using trajectories of  $N=32$ coupled oscillators, consisting of $n_s = 4, 000, 000$ data points, assuming the initial condition equivalent to giving the energy $\mathcal{E}_1  \approx 0.45$ to the first mode  ($k=1, A=10$). Note that the zero of the horizontal axis is set at $m =1$. (Inset) The same plot for  $m \in [60, 63]$ shows the curves corresponding to $\beta \in [2.4, 3]$.
        }
        \label{fig:reconstruction_err_k_1}
    \end{minipage}
    \hfill
    \begin{minipage}{0.48\textwidth}
        \centering
        \includegraphics[width=\linewidth]{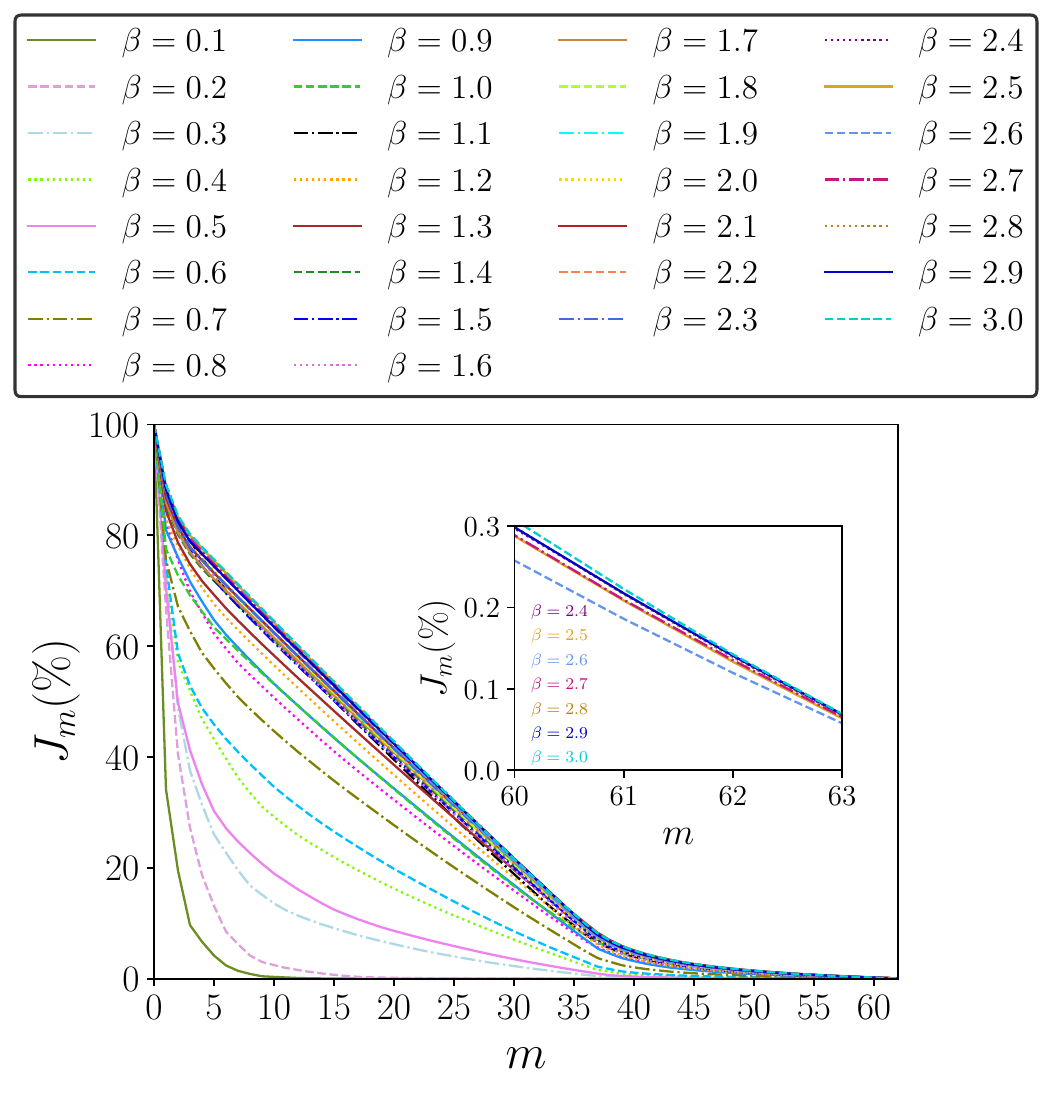}
        \caption{Reconstruction error $J_m$ in percentage ($\%$) as a function of the dimension $m$ of the best-fitting subspace $U$ for $\beta \in [0.1, 3]$, using trajectories of  $N=32$ coupled oscillators, consisting of $n_s = 4, 000, 000$ data points, assuming the initial condition equivalent to giving the energy $\mathcal{E}_1  \approx 1.8$ to the second mode  ($k=2, A=10$). Note that the zero of the horizontal axis is set at $m =1$. (Inset) The same plot for  $m \in [60, 63]$ shows the curves corresponding to $\beta \in [2.4, 3]$.}
        \label{fig:reconstruction_err_k_2}
    \end{minipage}
\end{figure*}

In Figs.\ref{fig:reconstruction_err_k_1} and \ref{fig:reconstruction_err_k_2}, the reconstruction error curves $J_m$ (in percentage), calculated using Eq.\ref{eq:rec_error_1}, are shown as functions of the dimension $m$ (that is, the number of principal components) of the best-fitting subspace, for the trajectory data corresponding to $k=1$ and $k=2$, respectively. 
When $k=1$, the curves form two families determined by the parameter $\beta$. One family emerges at small non-linearities, i.e., when $\beta \lesssim 1.1$, whose curves fall quickly, yielding very small intrinsic dimensions (see  Fig.~\ref{fig:family_of_curves} for better visualization of this family of curves).
The second family is formed by smoother curves that gradually decrease, starting from $\beta \gtrsim 1.1$. As a result, these curves yield larger intrinsic dimensions.  The origin of these different behaviors can be understood by examining the eigenvalues $\lambda_i$, contributing to Eq.~\ref{eq:rec_error_1}. When $\beta$ is small, only a few eigenvalues differ significantly from zero, as shown in Fig.~\ref{fig:eigenvalues_k_1_2}. For example, when $\beta=0.1$, $\lambda_1$ and $\lambda_2$ account for most of the preserved variance. In this case, the sum of the first two principal components PC1 and PC2, explains about $99\%$ of the data variability. Consequently, the curves in the first family diminish rapidly. In contrast, the curves of the other family originate from the contribution of a larger number of eigenvalues,  making them smoother and decaying more slowly.
In contrast, when $k=2$, all the reconstruction curves appear relatively smooth and decay slowly due to the smoother trends of their respective eigenvalues (see Fig.~\ref{fig:eigenvalues_k_1_2}). Furthermore, the insets of Figs.\ref{fig:reconstruction_err_k_1} and \ref{fig:reconstruction_err_k_2}, show how the curves corresponding to $\beta \in [2.4, 3]$, converge to zero linearly when $m$ approaches $n-1$.

\begin{figure}
\resizebox{0.50\textwidth}{!}{%
  \includegraphics{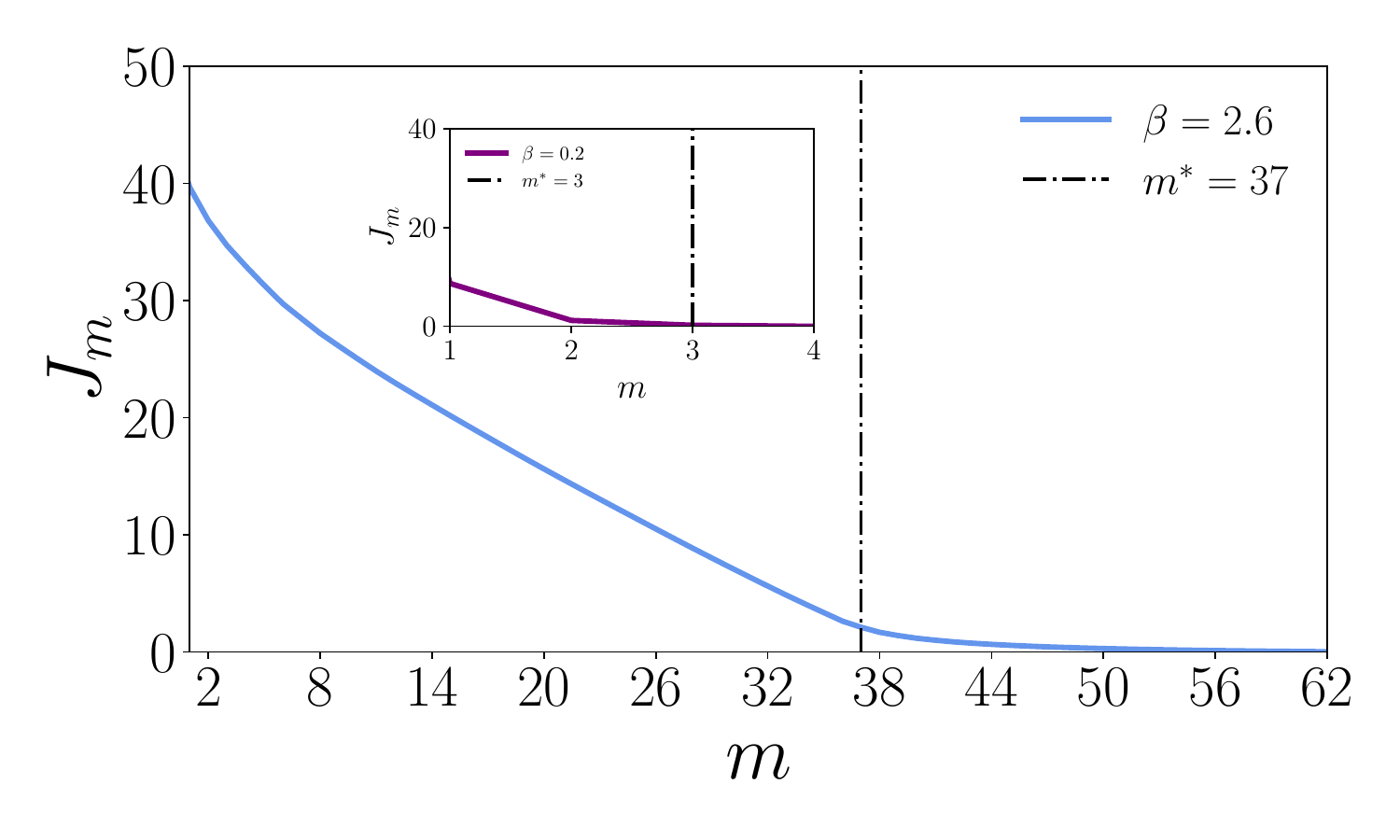}
}
\caption{KA with $s=1$ applied to the reconstruction error curves $J_m$ corresponding to $\beta=0.2$ (inset) $\beta=2.6$, obtained from trajectory data  ($n_s = 4,000,000$) with system size $N=32$, and the initial condition equivalent to giving $\mathcal{E}_1  \approx 0.45$ to the first  mode  ($k=1, A=10$). Each elbow point is declared at the intersection with the respective vertical line.}
\label{fig:kneedle}      
\end{figure}

Next, to automate the search for elbow points in the considered curves, we employ the Kneedle algorithm, a general-purpose knee detection method~\cite{ville2011}. This approach also helps mitigate the potential subjectivity and difficulty typically associated with this task~\cite{Ketchen1986}. Fig.~\ref{fig:kneedle} illustrates how KA works when applied to two specific reconstruction error curves $J_m$ ($k=1$), corresponding to $\beta=0.2$ (inset) and $\beta=2.6$, setting the parameter $s$, called sensitivity, to unity. Sensitivity measures the number of flat points in the curve before declaring the knee~\cite{ville2011}. 
In such a case, the algorithm finds the elbows, loosely assuming that they correspond to the points of maximum curvature. These points correspond to the intersection of the curves with the vertical lines, which yields $m^{\ast} =3$ and $m^{\ast} = 37$, respectively.
 These findings confirm what we would expect by visual inspection of the reconstruction curves, that is, $m^{\ast}$  increases with $\beta$. 

The Kaiser rule, used routinely in factor analysis, states that only the principal components with $\lambda_i \geq 1$ should be retained~\cite{Kaiser1960}. Based on simulation studies,  Jolliffe later suggested that, in the context of PCA, a more reasonable threshold is given by $\lambda_i \geq 0.7$~\cite{jolliffe2002pca}. In the following, we shall adopt the Jolliffe ansatz. Finally, the participation ratio is defined as~\cite{RECANATESI2022}
\begin{equation}\label{eq:participation}
D_{\rm PR} = \frac{\left( \sum\limits_{i=1}^{n} \lambda_i \right)^2}{\sum\limits_{i=1}^{n} \lambda_i^2} \, .
\end{equation}

Note that Eq.~\ref{eq:participation} can also be written in terms of the traces of matrices $S$ (Eq.~\ref{eq:covariance}) and $S^{2}$, respectively, as $D_{\rm PR} = \left( \operatorname{Tr}(S) \right)^2/\operatorname{Tr}(S^2)$. The $D_{\rm PR}$ measures the concentration of the eigenvalue distribution, which yields the number of PCs that capture most of the variance~\cite{RECANATESI2022}.

It is important to note that, due to their heuristic nature, the above methods generally cannot guarantee optimal results; therefore, their findings should be assessed using other nonlinear approaches, as discussed in Sec.~\ref{conclusions}.

\begin{figure*}
    \centering
    \begin{minipage}{0.48\textwidth}
        \centering
        \includegraphics[width=\linewidth]{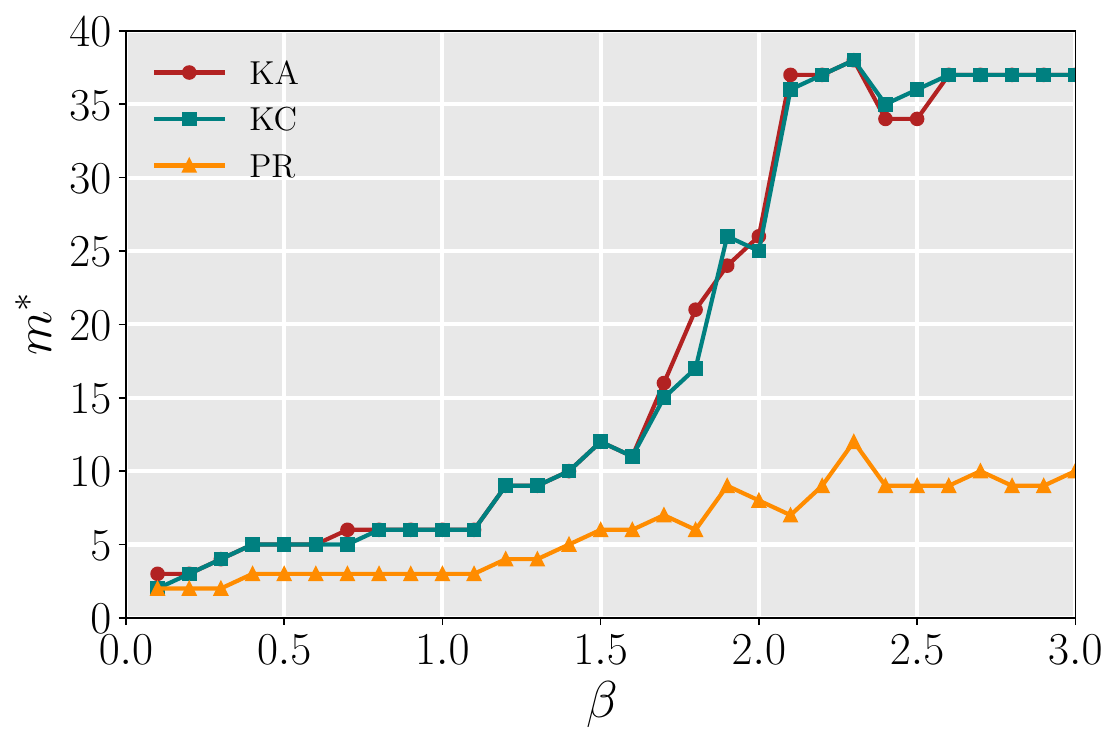}
        \caption{Estimated intrinsic dimension $m^{\ast}$ as a function of $\beta$, obtained using KA, KC, and PR. Each trajectory dataset contains $n_s = 4,000,000$ data points for each $\beta$. The initial condition corresponds to exciting the first mode with energy $\mathcal{E}_1 \approx 0.45$ (i.e., $k = 1$, $A = 10$).}
        \label{fig:dimensions_k_1}
    \end{minipage}
    \hfill
    \begin{minipage}{0.48\textwidth}
        \centering
        \includegraphics[width=\linewidth]{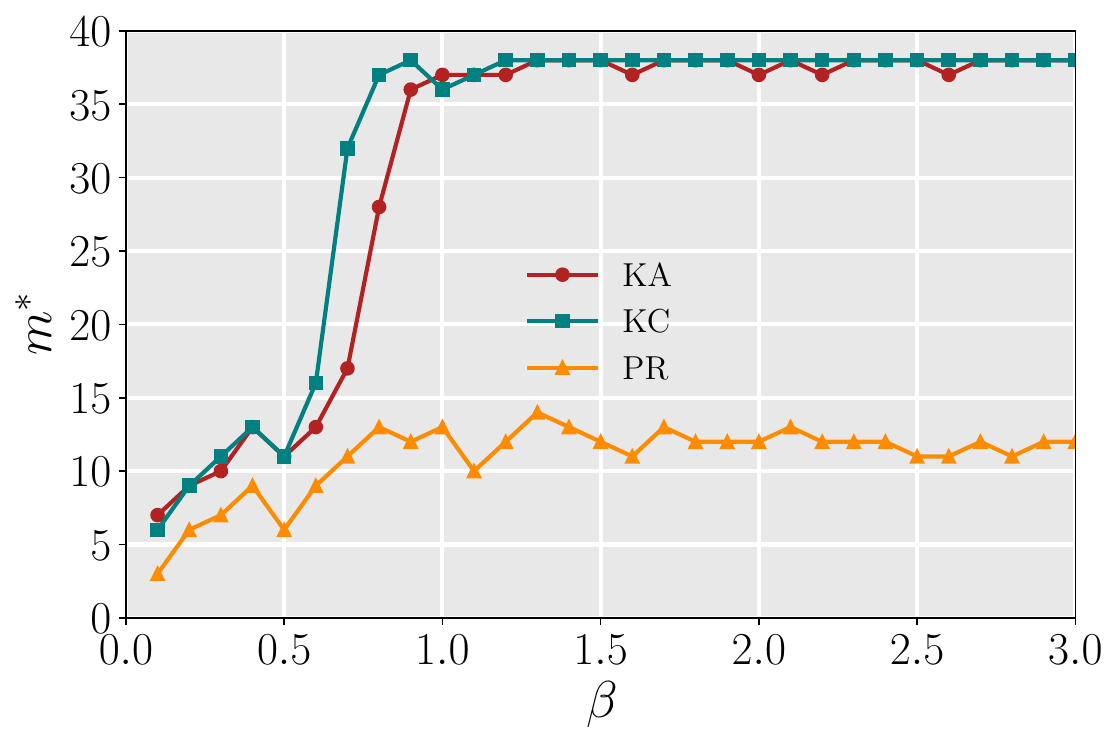}
        \caption{Estimated intrinsic dimension $m^{\ast}$ as a function of $\beta$, obtained using KA, KC, and PR. Each trajectory dataset contains $n_s = 4,000,000$ data points for each $\beta$. The initial condition corresponds to exciting the second mode with energy $\mathcal{E}_1 \approx 1.85$ (i.e., $k = 2$, $A = 10$).}
        \label{fig:dimensions_k_2}
    \end{minipage}
\end{figure*}

To begin with, Fig.\ref{fig:dimensions_k_1} shows the intrinsic dimension $m^{\ast}$ as a function of $\beta$ for the case $k = 1$, estimated using the Kneedle algorithm (circle symbols), the Kaiser criterion (square symbols), and the participation ratio (triangle symbols). In general, the respective curves exhibit a monotonic trend with increasing $\beta$. In particular, we note that only KA and KC show close numerical agreement throughout the range. In the weakly nonlinear regime ($\beta \lesssim 1.1$), KA and KC estimate $m^{\ast} = 3$–$6$, while PR yields lower values of $m^{\ast} = 2$–$3$, in good agreement with the multi-chart approach proposed by Yu et al.~\cite{Yu2025, Hauberg2025}. However, beyond this point, a clear discrepancy emerges between the methods. The PR curve increases monotonically but very slowly, reaching $m^{\ast} = 10$ as $\beta \rightarrow 3$. In contrast, the KA and KC curves exhibit a sharp increase and quickly converge to $m^{\ast} = 36$–$37$ from $\beta \gtrsim 2.1$. In this regard, we argue that only the KA and KC methods likely capture the sudden changes that occur in the dynamics of the system, as $\beta$ increases. Our previous observation is based on the patterns observed through the Poincar\'{e}  maps~\cite{TUCKER2002127} for the problem at hand. Accordingly, the Poincar\'{e}  maps indicate that the regular patterns associated with quasi-periodic motion (for $\beta \lesssim 1.1$) gradually disappear as $\beta$ increases. In their place, a clear emergence of randomness is observed, strongly suggesting that the system is transitioning toward a chaotic regime~\cite{giordano2006}. 
Furthermore, it is also plausible that the substantial changes in dimensionality, observed just after the recurrent motion regime, are driven by symmetry breaking, enabling the first mode to efficiently exchange energy with other modes. 

On the other hand, the high dimensionality of the trajectory data observed for $\beta \gtrsim 2.1$ corresponds to a regime in which the system approaches thermal equilibrium. In this case, we found that at the end of the simulations, the excited initial mode ($k=1$) has shared nearly all of its energy with the other modes. As a result, the mode energies $E_i$ tend to satisfy $E_i \approx \epsilon_1$. The previous finding is further confirmed by doubling the simulation time, achieved by increasing the integration step to $h = 0.1$ according to Ref.~\onlinecite{Livi1985}.

Finally, we focus on the trajectory data corresponding to the initial condition $k = 2$ ($A = 10$). In this case, a higher energy density $\epsilon_2 \approx 4$ (see Sec.\ref{fput_facts} for details) leads to stronger nonlinear effects in the dynamics. As a result, energy recurrences are observed only at $\beta = 0.1$ (see Fig.\ref{fig:energy_modes_0.1_k_2}). For higher values of $\beta$, the second mode begins to efficiently share its energy with other modes after a transient period, which becomes shorter as $\beta$ increases, as illustrated in Figs.\ref{fig:energy_modes_0.2_k_2},\ref{fig:energy_modes_0.3_k_2}, and~\ref{fig:energy_modes_0.4_k_2}.
The corresponding KA, KC, and PR curves as functions of $\beta$ are shown in Fig.~\ref{fig:dimensions_k_2}. As in the previous case, the curves exhibit a clear monotonic trend for $\beta \lesssim 1$, after which they rapidly converge to  $m^{\ast} = 11$–$12$, and  $m^{\ast} = 37$–$38$, according to PR and KA, and KC, respectively. This behavior confirms that the high intrinsic dimensionality of the data primarily arises from the system's strong nonlinearity. Notably, PR yields $m^{\ast} = 3$ at $\beta = 0.1$, a reasonable value that supports the earlier observation that quasi-periodic motion occurs on a low-dimensional Riemannian manifold. However, as before, the PR curve lacks the dramatic changes observed in KA and KC curves as $\beta$ increases.
Furthermore, the KA and KC curves quickly converge as $\beta$ increases, thus estimating $m^{\ast} = 38$. This finding appears to characterize the approach to equilibrium of the $\beta$ model with $N = 32$, consistent with the previous case of $k = 1$. In this regard, it is worth noting that the high-dimensional trajectories will eventually occur even for weak nonlinearities, as the system slowly approaches equilibrium~\cite{Livi1985, Ford1970}.

We conclude by noting that most of the results presented here should be regarded as crude approximations of the true intrinsic dimensionality of the data, owing to the inherent limitations of linear approaches such as PCA.
In Sec.~\ref{conclusions}, we outline potential strategies for improving upon principal component analysis and discuss possible directions for future research.

\section{\label{conclusions} Conclusion}

In this exploratory work, we presented a data-driven approach based on principal component analysis to investigate the rich phenomenology of the FPUT $\beta$ model, using full trajectory data accurately obtained using the Verlet algorithm.  Despite the limitation of such a linear approach, some of which are addressed using $t$-SNE, we find a crucial relationship between the intrinsic dimensionality of the trajectories and the nonlinearity strength of the model. PCA suggests that for weak nonlinearity, where energy recurrences are observed, the trajectories lie on or near a two- or three-dimensional hyperplane. This finding is in numerical agreement with results obtained using the multi-chart flows method recently proposed by Yu et al.~\cite{Yu2025}. However, only the latter can correctly prove that the periodic motion of the system takes place on a low-dimensional Riemannian manifold. In contrast, high intrinsic dimensionality is characteristic of stronger nonlinearities, where energy is efficiently exchanged among modes, enabling the system to reach thermal equilibrium.

Similar studies using alternative manifold learning algorithms, such as kernel PCA~\cite{scholkopf1998}, the approach based on the multi-chart flows~\cite{Yu2025}, and neural network architectures like autoencoders~\cite{Wehmeyer2018, otto2019, agostini2020, glielmo2021, Bonheme2022}, are very likely to provide a more accurate estimate of the correct data dimensionality, which remains beyond the reach of the principal component analysis.

Here, we focus on a minimal FPUT $\beta$ model with $N = 32$. For future research, it would be valuable to investigate how system size $N$ influences data dimensionality. Furthermore, it would be of interest to apply a similar data-driven analysis to other variants of the FPUT model, such as the $\alpha$ model and the combined $\alpha + \beta$ model.

Finally, there is strong evidence supporting the existence of a Riemannian manifold on which the trajectory lies in the weakly nonlinear regime. This manifold, and its potential change with increasing nonlinearity, could be effectively explored using topological data analysis (TDA)~\cite{Carlsson2009, Munch_2017, Otter2017, chazal2021, Papillon2025} or geometric data analysis (GDA)~\cite{hickokthesis2023}. For example, persistent homology, a tool from TDA, can quantify topological features of the data such as the number of connected components, holes, and higher-dimensional voids. Similarly, GDA offers insights by analyzing geometric invariants of the manifold, such as its curvature~\cite{hickok2023}. In particular, using TDA and GDA could make it possible to investigate whether the symmetry breaking observed in the $\beta$ model is a consequence of changes in the topological and geometric features of the underlying Riemannian manifold.

\begin{acknowledgments}
\noindent The author thanks Hanlin Yu, S\o{}ren Hauberg, and Georgios Arvanitidis for analyzing a dataset using their Riemannian manifold learning approach. The author also acknowledges Angelo Vulpiani for discussions on Hamiltonian systems and the FPUT model. Furthermore, the author is grateful to Dmitry Kobak for helpful correspondence during the preparation of this article and for suggesting the use of the Python library \textsf{openTSNE}~\cite{policar2024}, and to Giancarlo Benettin for correspondence regarding the FPUT model. Finally, the author is indebted to Jack Dongarra and David Keyes for sharing their recent review paper~\cite{Dongarra2024}. Code execution for this project was performed using Google Colaboratory.\\
\end{acknowledgments}

\section*{Data Availability Statement}

The dataset used in this study is available at Zenodo~\cite{marchetti_2025_a, marchetti_2025_b}.

\appendix

\section{\label{fput_facts} Initial Conditions and Simulations}

\begin{figure}[h]
\resizebox{0.50\textwidth}{!}{%
  \includegraphics{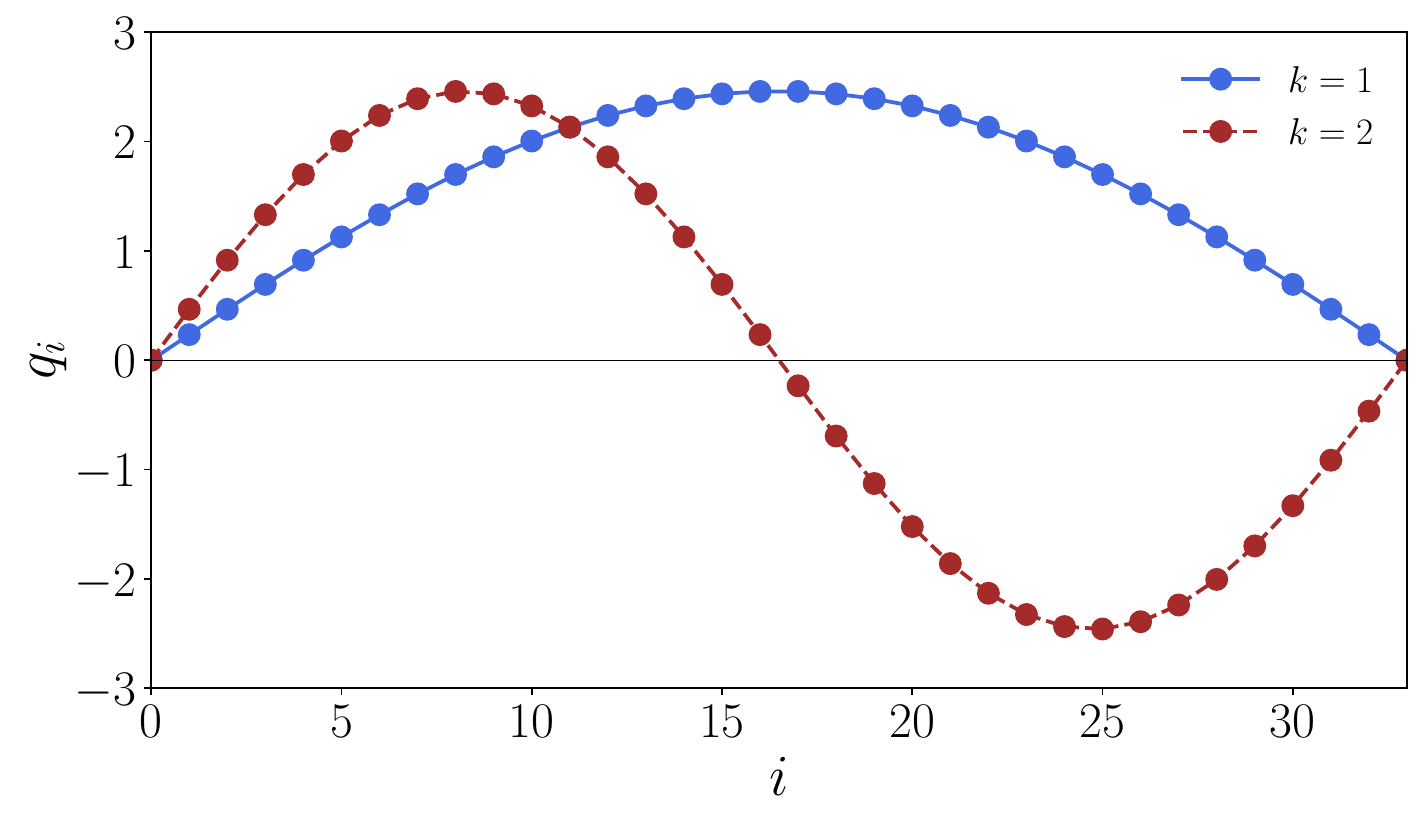}
}
\caption{The coordinates $q_i$ ($i=1, \cdots, 32$) at time $t=0$ according to Eq.~\ref{eq:initial_cond}, assuming  to initially exciting the first mode $k=1$ (solid line) or the second mode $k=2$ (dashed line). In both cases $A=10$.}
\label{fig:initial_modes}      
\end{figure}

The FPUT trajectories under scrutiny start either from
the initially excited first mode or from the initially excited second mode, computed, setting $k=1, A=10$ and $k=2, A=10$ in Eq.~\ref{eq:initial_cond}, respectively. These initial conditions correspond to initially displacing the coordinates $q_i$ ($i=1, \cdots, 32$) as depicted in Fig.~\ref{fig:initial_modes} with a solid ($k=1, A=10$), and a dashed line ($k=2, A=10$). The initial conditions for the variables $p_i$ are $p_i =0$ with $i =1, \cdots, 32$. 

Accordingly, the energy of the linear system takes the values $\mathcal{E}_1 \approx 0.45$ and $\mathcal{E}_2  \approx 1.8$ when $k=1, A=10$ and $k=2, A=10$, respectively. 

Here, it is worth noting that assuming weak nonlinearity, i.e. $\beta \approx 0$, the system's energy density is $\epsilon_1 \approx 14 \times 10^{-3} $ and $\epsilon_2 \approx 56 \times 10^{-3}$, for $k=1$ and $k=2$, respectively. As a result, $\epsilon_2 \approx 4 \epsilon_1$. Therefore, for a given small value of $\beta$, the model dynamics with the initial condition $k=2$ ($A=10$) is subject to a stronger nonlinearity compared to the case with the initial condition $k=1$ ($A=10$).

\begin{figure*}
    \centering
    \begin{minipage}{0.48\textwidth}
        \centering
        \includegraphics[width=\linewidth]{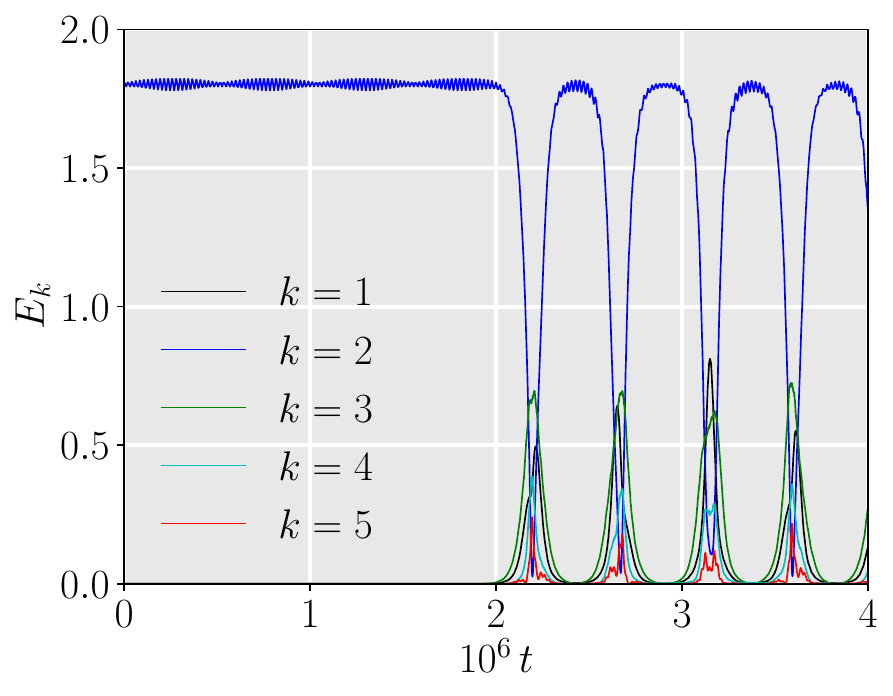}
        \caption{The energy $E_k$  of modes with $k=1, 2, 3, 4, 5$ as a function of the  time  $t$  for $\beta$ model with $\beta=0.1$, assuming $N=32$. The system's equations of motion were numerically integrated with step size $h=0.05$.  The initial condition is set to provide the energy $\mathcal{E} \approx 1.8$ to the second normal mode ($k=2, A=10$).}
        \label{fig:energy_modes_0.1_k_2}
    \end{minipage}
    \hfill
    \begin{minipage}{0.48\textwidth}
        \centering
        \includegraphics[width=\linewidth]{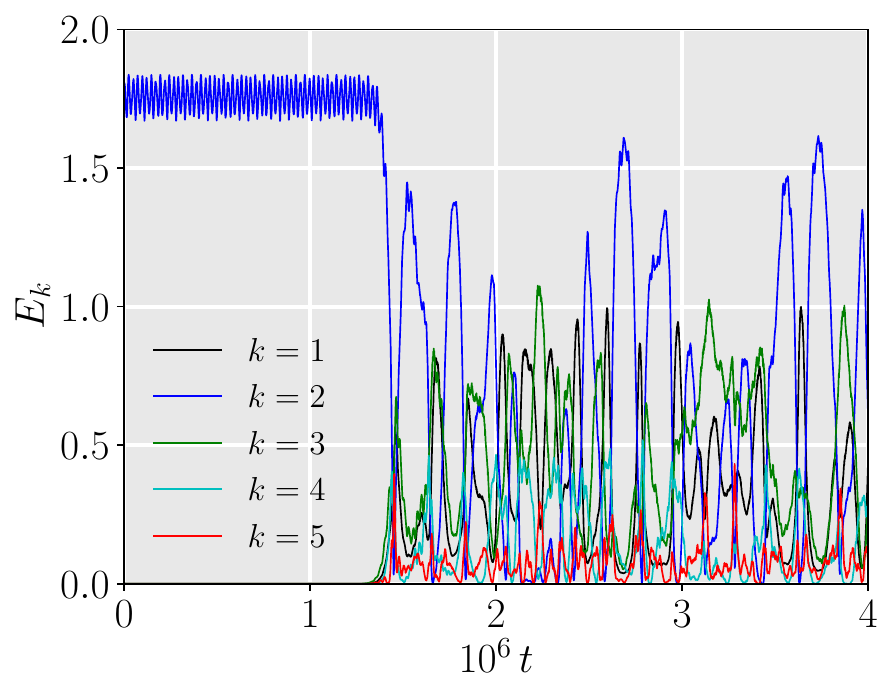}
        \caption{The energy $E_k$  of modes with $k=1, 2, 3, 4, 5$ as a function of the  time  $t$  for $\beta$ model with $\beta=0.2$, assuming $N=32$. The system's equations of motion were numerically integrated with step size $h=0.05$.  The initial condition is set to provide the energy $\mathcal{E} \approx 1.8$ to the second normal mode ($k=2, A=10$).}
        \label{fig:energy_modes_0.2_k_2}
    \end{minipage}
\end{figure*}

\begin{figure*}
    \centering
    \begin{minipage}{0.48\textwidth}
        \centering
        \includegraphics[width=\linewidth]{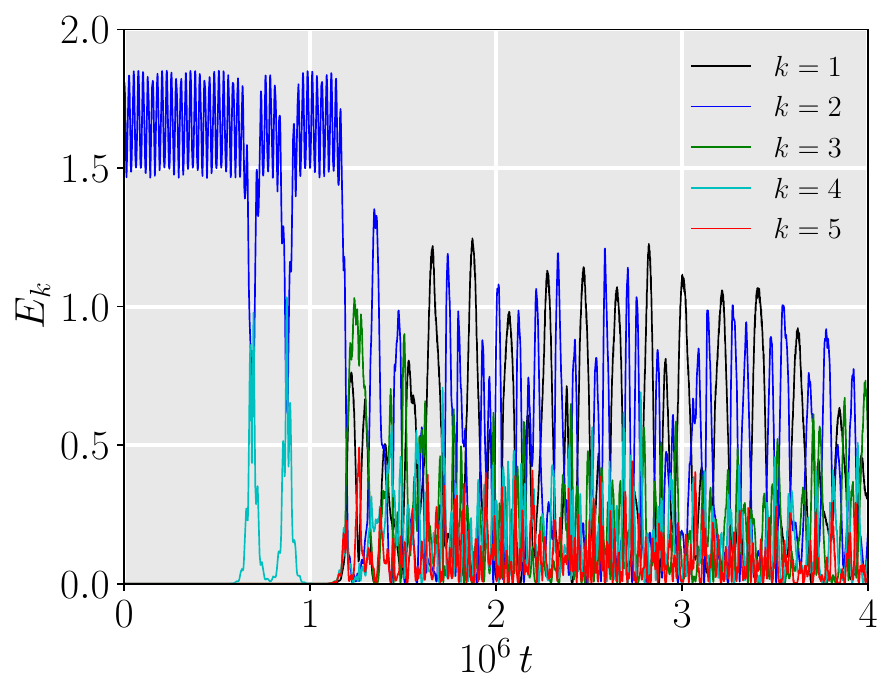}
        \caption{The energy $E_k$  of modes with $k=1, 2, 3, 4, 5$ as a function of the  time  $t$  for $\beta$ model with $\beta=0.3$, assuming $N=32$. The system's equations of motion were numerically integrated with step size $h=0.05$.  The initial condition is set to provide the energy $\mathcal{E} \approx 1.8$ to the second normal mode ($k=2, A=10$).}
        \label{fig:energy_modes_0.3_k_2}
    \end{minipage}
    \hfill
    \begin{minipage}{0.48\textwidth}
        \centering
        \includegraphics[width=\linewidth]{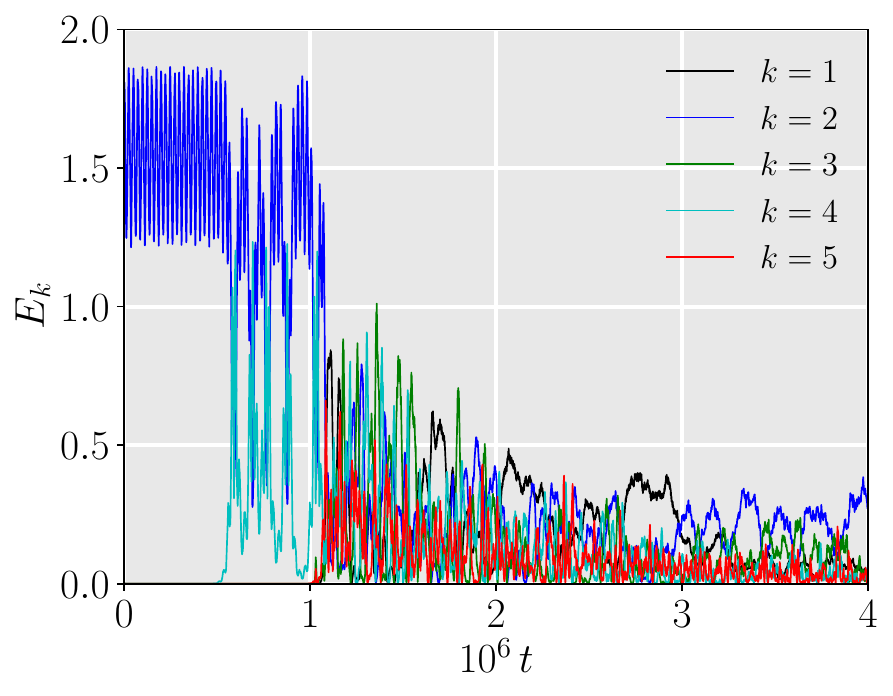}
        \caption{The energy $E_k$  of modes with $k=1, 2, 3, 4, 5$ as a function of the  time  $t$  for $\beta$ model with $\beta=0.4$, assuming $N=32$. The system's equations of motion were numerically integrated with step size $h=0.05$.  The initial condition is set to provide the energy $\mathcal{E} \approx 1.8$ to the second normal mode ($k=2, A=10$).}
        \label{fig:energy_modes_0.4_k_2}
    \end{minipage}
\end{figure*}

In Figs.~\ref{fig:energy_modes_0.1_k_2}, ~\ref{fig:energy_modes_0.2_k_2}, ~\ref{fig:energy_modes_0.3_k_2} and \ref{fig:energy_modes_0.4_k_2}, the time-evolution of energy of the first five normal modes as a function of time $t$, for $\beta=0.1, 0.2, 0.3, 0.4$, assuming the initial condition with $k=2$ ($A=10$). We note that energy recurrences now occur only for $\beta=0.1$, while the initially excited mode $E_1$ begins to efficiently share its energy with the others, after an initial transient time, which becomes shorter as $\beta$ increases. These findings illustrate that stronger nonlinearity is present when $k=2$, due to the higher energy density.

\subsection{\label{pca_facts} PCA Results}

\begin{figure}
\resizebox{0.50\textwidth}{!}{%
  \includegraphics{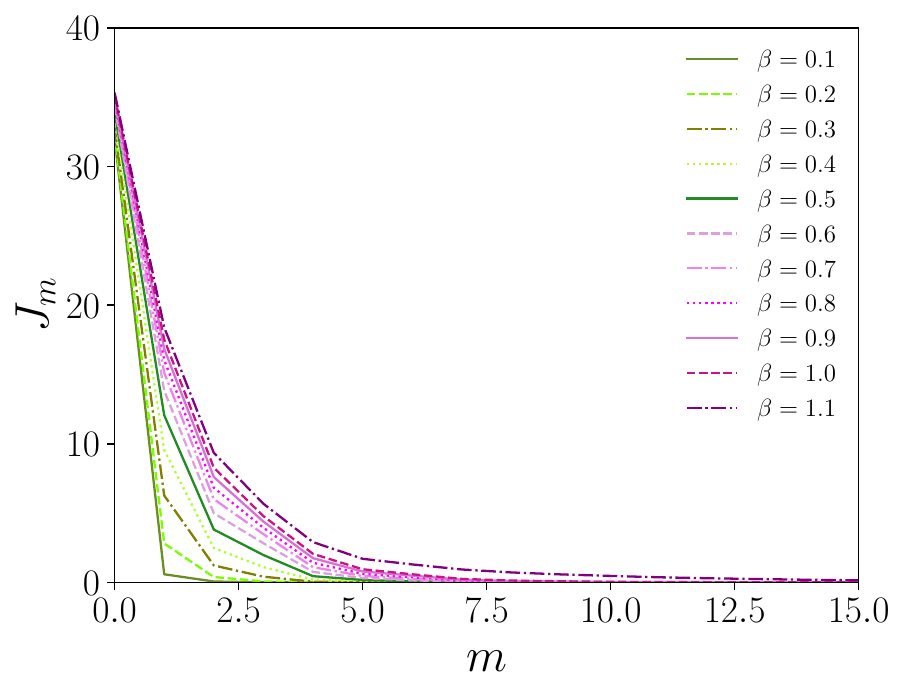}
}
\caption{Reconstruction error $J_m$ as a function of the dimension $m$ of the best-fitting subspace $U$ for $\beta \in [0.1, 1.1]$, using trajectories of  $N=32$ coupled oscillators, consisting of $n_s = 4, 000, 000$ data points, assuming the initial condition equivalent to giving the energy $\mathcal{E}_1  \approx 0.45$ to the first mode  ($k=1, A=10$). Note that the zero of the horizontal axis is set at $m =1$.}
\label{fig:family_of_curves}      
\end{figure}

In Fig.~\ref{fig:family_of_curves} only the reconstruction error curves $J_m$ for $\beta \in [0.1, 1.1]$ ($k=1$, $A=10$) are shown as functions $m$ for better visualization.

\begin{figure*}[htb]
   \includegraphics[width=\textwidth]{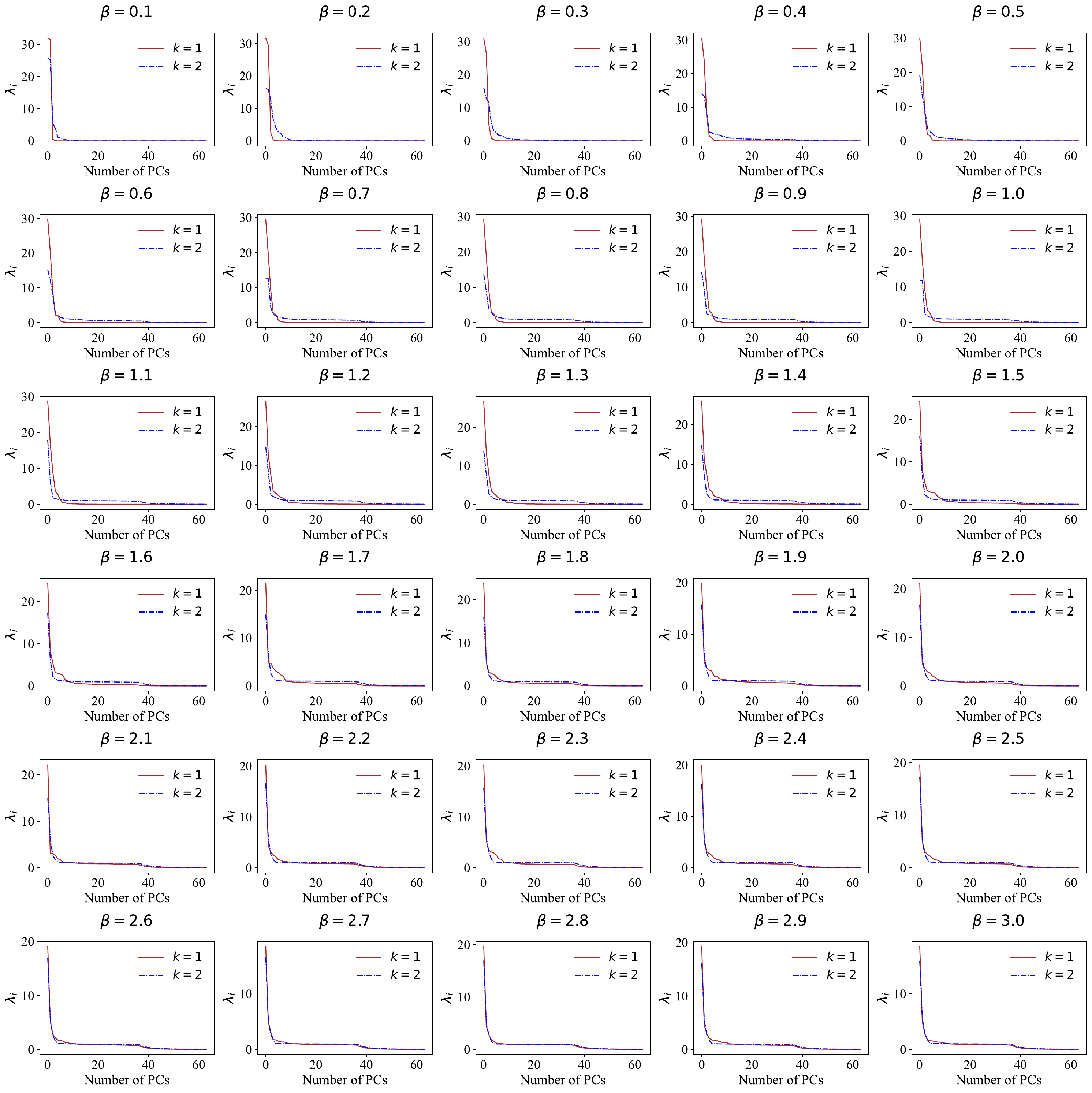}
\caption{Eigenvalues  $\lambda_i$ as functions of the number of the principal components PCs according to SVD applied to data from the entire trajectories ($n_s=4, 000, 000$), assuming the system size $N=32$ and $\beta \in [0.1, 3]$. The initial conditions of the trajectory data correspond  to initially exciting the first mode  (solid curves) or the second mode (dashdot curves), assuming $A=10$. Note that the zero of the horizontal axis is set at the first principal component.}
\label{fig:eigenvalues_k_1_2}      
\end{figure*}

In Fig.~\ref{fig:eigenvalues_k_1_2} the eigenvalues  $\lambda_i$ ($i = 1, 2, \dots, 64)$ of the correlation matrix, obtained by singular value decomposition of the data matrix from the trajectory data ($n_s = 4, 000, 000$), with initial conditions $k=1$ and $k_2$, respectively, are shown as functions of the number of principal components, for each value of $\beta$ under scrutiny. It should be noted that for $k=1$, PC1 + PC2 together
account for between  $71\%$ and  $99\%$ of the variance preserved when $\beta \in [0.1, 1.1]$. In contrast, for $k=2$,  the explained variance exceeds $70\%$ only at $\beta = 0.1$, where it reaches approximately $79\%$.

\bibliographystyle{unsrturl} 
\bibliography{aipsamp}
\end{document}